\documentclass[conference]{IEEEtran}
\IEEEoverridecommandlockouts
\usepackage{cite}
\usepackage{amsmath,amssymb,amsfonts}
\usepackage{hyperref}
\usepackage{algorithmic}
\usepackage{algorithm}
\usepackage{graphicx}
\usepackage{textcomp}
\usepackage{multicol,multirow}
\usepackage[center]{caption}
\usepackage{xcolor}
\def\BibTeX{{\rm B\kern-.05em{\sc i\kern-.025em b}\kern-.08em
    T\kern-.1667em\lower.7ex\hbox{E}\kern-.125emX}}

\makeatletter

\begin{document}

\title{A New Lightweight Hybrid Graph Convolutional Neural Network - CNN Scheme for Scene Classification using Object Detection Inference
}

\author{
    \IEEEauthorblockN{\hspace{-1cm}1\textsuperscript{st} Ayman Beghdadi}
    \IEEEauthorblockA{\hspace{-1cm}\textit{Paris-Saclay University} \\
    \hspace{-1cm}Evry, France \\
    \hspace{-1cm}aymanaymar.beghdadi@univ-evry.fr}
    \and
    \IEEEauthorblockN{\hspace{-1cm}2\textsuperscript{nd} Azeddine Beghdadi}
    \IEEEauthorblockA{\hspace{-1cm}\textit{Sorbonne Paris Nord University} \\
    \hspace{-1cm}Villetaneuse, France\\
    \hspace{-1cm}azeddine.beghdadi@univ-paris13.fr}
    \and
    \IEEEauthorblockN{\hspace{-1cm}3\textsuperscript{rd} Mohib Ullah}
    \IEEEauthorblockA{\hspace{-1cm}\textit{NTNU University} \\
    \hspace{-1cm}Gjovik, Norway\\
    \hspace{-1cm}mohib.ullah@ntnu.no}
    \and  
    \hspace{4cm} 
    \IEEEauthorblockN{4\textsuperscript{th} Faouzi Alaya Cheikh}
    \IEEEauthorblockA{\hspace{4cm}\textit{NTNU University} \\
    \hspace{4cm}
    Gjovik, Norway\\
    \hspace{4cm}
    faouzi.cheikh@ntnu.no}
    \and
    \IEEEauthorblockN{\hspace{-1cm}5\textsuperscript{th} Malik Mallem}
    \IEEEauthorblockA{\hspace{-1cm}\textit{Paris-Saclay University} \\
    \hspace{-1cm}Evry, France\\
    \hspace{-1cm}malik.mallem@univ-evry.fr}
}

\maketitle
\IEEEaftertitletext{\vspace{-3\baselineskip}}
\begin{abstract}

Scene understanding plays an important role in several high-level computer vision applications, such as autonomous vehicles, intelligent video surveillance, or robotics. However, too few solutions have been proposed for indoor/outdoor scene classification to ensure scene context adaptability for computer vision frameworks. We propose the first Lightweight Hybrid Graph Convolutional Neural Network (LH-GCNN)-CNN framework as an add-on to object detection models. The proposed approach uses the output of the CNN object detection model to predict the observed scene type by generating a coherent GCNN representing the semantic and geometric content of the observed scene. This new method, applied to natural scenes, achieves an efficiency of over 90\% for scene classification in a COCO-derived dataset containing a large number of different scenes, while requiring fewer parameters than traditional CNN methods.
For the benefit of the scientific community, we will make the source code publicly available: \href{https://github.com/Aymanbegh/Hybrid-GCNN-CNN}{https://github.com/Aymanbegh/Hybrid-GCNN-CNN}.
\end{abstract}

\begin{IEEEkeywords}
Deep learning, Graph neural network, Indoor/outdoor, Scene classification, Scene understanding
\end{IEEEkeywords}

\section{Introduction}
\label{sec:intr}

Nowadays, mobile and robotic systems have achieved a considerable level of autonomy thanks to the development of new computer vision methods based on deep learning.
Indeed, over the last decade, the evolution of deep learning techniques has enabled us to perform increasingly complex and diversified tasks such as object detection and segmentation \cite{joseph2021towards}, object and action recognition \cite{hussein2019timeception}.
Autonomous systems are capable of performing complex tasks in unfamiliar environments with good reliability.
This has enabled us to offer a solution for a wide range of fields.

\begin{figure}[!t]
    \centering
    \includegraphics[width=0.48\textwidth]{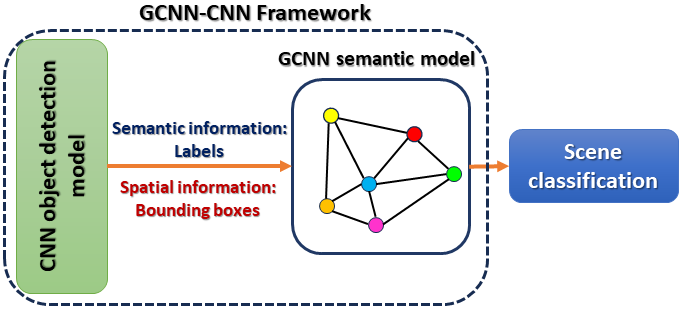}
    \caption{LH-GCNN-CNN Framework.}
    \label{fig:intro}
    \vspace{-1.5em}
\end{figure}

In the context of autonomous navigation based on computer vision, i.e. visual SLAM (vSLAM), most existing methods \cite{nardi2020long,ehlers2020map} are dedicated to a specific type of environment, indoor or outdoor.
Most visual SLAM frameworks incorporate object detection and/or segmentation models, enabling scene content analysis for complex tasks.
In particular, the need for vSLAM methods to be robust to real environments containing dynamic objects has led to a rethinking of these methods.
Indeed, the complexity of the tasks to be performed and of the environments in which autonomous systems operate has led us to impose strong assumptions on the system.

In general, dynamic vSLAM methods \cite{yu2018ds,ayman2023dam} include assumptions linked to the application framework to optimize the efficiency of the complex tasks to be carried out.
These methods consider assumptions about the dynamic state of certain objects as a function of the type of environment (indoor/outdoor) to not consider features from dynamic areas.
In most cases, these assumptions are immutable and for a single specific setting, which is particularly restrictive.

Consequently, we propose a solution for dynamically adjusting these assumptions according to the type of scene (indoor/outdoor) by using an hybrid GCNN-CNN framework as shown in Fig.\ref{fig:intro}.
To this end, we have developed a GCNN model that performs scene classification (indoor/outdoor) by exploiting spatial and semantic information obtained from CNN object detection/segmentation models.
This approach offers an efficient solution for extracting scene features and context relationships by exploiting the advantages of graphs as an efficient tool for representing complex data.
To the best of our knowledge, this work is the first to address this issue by combining CNN and GCNN models for predicting scene type of natural scenes.
\newline

The proposed  model can be integrated as a framework within the well-known YOLACT \cite{bolya2019yolact} object detection or segmentation model. It can also be extended to any model that outputs labels and bounding boxes for detected objects.
Furthermore, the introduced GCNN model can be added as a network header to any object detection/segmentation model, making it easier to use in various computer vision applications.
The main contributions of this work are:
\begin{itemize}
\item A Lightweight Hybrid GCNN-CNN framework for scene classification is proposed, with predictive accuracy of over 90\% and a much lower number of parameters than CNN models.
\item A GCNN-based solution for non-satellite image scene classification is introduced for the first time and validated through extensive experiments on real data.
\item An efficient graph construction strategy exploiting both semantic and spatial information of objects, and their distribution in the scene is proposed.
\item A new GCNN-based framework incorporating the semantic contents and the spatial distribution of objects for better extraction of scene features and visual information context, i.e. objects relationship, in order to learn intrinsic attributes of the observed scenes is proposed.
\item A complete and flexible open-source code that can be easily integrated with any object detection/segmentation model for solving various computer vision problems is provided.
\end{itemize}
The paper is organized as follows. Section \ref{sec:relat} summarizes the related literature. 
Section \ref{sec:meth} is devoted to detail our method. Results are presented and discussed in section \ref{sec:expe}.
Conclusions and perspectives are provided in section \ref{sec:conc}.

\section{Related works}
\label{sec:relat}
The classification of scenes is a hot research topic, dating back to at least the 90s \cite{yu1995scenic}. Despite the number of works dedicated to this subject, whether for indoor or outdoor scene classification or other contexts, it remains an open research problem.
In what follows, we will limit ourselves to a few published works that we feel are representative of the three decades of research on indoor-outdoor scene classification.
\subsection{Indoor/outdoor scene classification}
With the advent of deep learning techniques, great progress has been made in the field of image classification, and several survey papers have highlighted this \cite{tong2017review, cheng2020remote}. In the case of indoor/outdoor scene classification (binary decision), although this appears simpler than image classification in the broad sense, there are cases where the classification error is not negligible. Indeed, recognizing and classifying objects in an image may be less difficult than classifying indoor and outdoor scenes because of the difficulty in discerning between perceptually and physically similar visual information that may exist in both indoor and outdoor scenes.
Given the large amount of work in the field and the trend over the last ten years, it seems more appropriate to classify the methods into two categories: traditional and deep learning-based. 

\textbf{Traditional approaches}
This category includes Bayesian framework approaches \cite{kane2004bayesian}, clustering methods based on low-level features of bag of words \cite{gokalp2007scene}, conventional learning-based methods such as SVM-based approaches \cite{serrano2002computationally}, random Forest classifiers \cite{bosch2007image}  and other approaches such as those based on the interesting concept of semantic typicality introduced in by Vogel and Schiele in \cite{vogel2004semantic}.
\begin{figure*}[!h]
  \centering
    \includegraphics[width=0.99\textwidth]{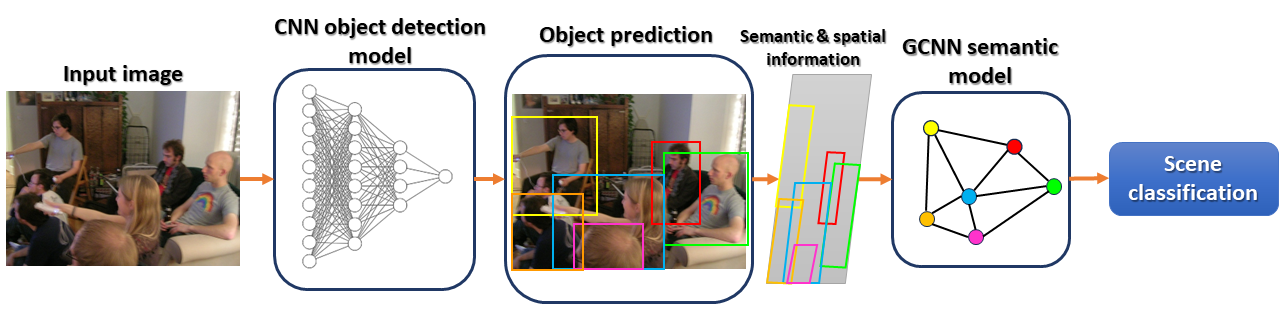}
  \caption{Architecture of the proposed method exploiting semantic and spatial information from the object detection model.}
  \label{fig:archi}
  \vspace{-1em}
\end{figure*}

\textbf{Deep learning based methods}
With the introduction of deep neural learning approaches, real progress has been made in recent years in solving scene classification problems, in general, \cite{lin2023benefits} and especially in the case of Remote-Sensing  \cite{wang2023remote}. But in the case of indoor/outdoor scene classification, despite the progress made, the problem still remains open due to its complexity \cite{lin2023benefits}. The other difficulty is the fact that the performance of deep learning-based methods depends on the quality of the databases used in training. It is also worth noting the lack of sufficient databases of labelled indoor-outdoor scenes available to the public, as highlighted in \cite{tong2017review}. 

\subsection{GCNN-based scene classification}
In this work, we focus on a few recent and representative state-of-the-art (SOTA) methods based on GCN architecture.
Liang et al. \cite{liang2020deep} propose a method that learns discriminative features of the scene by using the GCNN-CNN model. This approach extracts global-based visual features of scenes through a CNN architecture and then constructs graphs to learn the object-based location features. This method seems efficient for aerial and satellite views but unsuitable for common natural scenes due to its inability to exploit the spatial distribution of objects and their characteristic information.

Li et al. \cite{li2021robust} construct a new remote sensing knowledge graph (RSKG) to recognize unseen remote sensing image scenes. It introduces a generation of Semantic Representation of scene categories used by a deep alignment network (DAN) to exploit cross-modal matching between visual features and semantic representations. However, this approach does not take into account the spatial information of objects or their dimensions when creating graphs, which also makes it unsuitable for natural scenes.

In \cite{xu2021deep}, the proposed method exploits the benefits of GCN to better extract potential context relationships between objects in the observed scene. 
The proposed framework exploits only semantic scene information and spatial object distribution. However, this approach does not consider the object's intrinsic spatial information, making it ineffective for terrestrial images. 
It should be noted that most indoor/outdoor scene classification methods do not jointly incorporate information on the size of objects, as well as proximity characteristics and links between the various attributes of the image components observed in the learning process design. 

Moreover, these methods are based on a graph construction strategy that does not incorporate the most relevant intrinsic scene attributes, such as spatial and context relationships between the objects within the scene. Indeed, the generated graphs have all their nodes connected to each other, with no distinction between possible objects' relationships.
Our approach overcomes all these limitations by providing a real solution to the problem of non-satellite image scene classification.


\section{Method}
\label{sec:meth}
The overall architecture of our new Lightweight Hybrid Graph Convolutional Neural Network - CNN Scheme (LH-G2CNN) is illustrated in Fig.\ref{fig:archi}. The proposed model consists of a CNN object detection model that conveys semantic and spatial information to a GCNN model for boosting the scene classification process.

\subsection{Problem formulation}

The proposed method of scene classification by means of object detection inference aims to produce a learning-based classifier by using semantic information and spatial consistency of the scene.
First, the output prediction of the object detection inference provides a set of $n$ labels $L$ and bounding boxes $B$ describing the contents of the scene. Then, we use these characterizing features to produce a scene description exploiting the semantic and spatial scene contents and the spatial relation between the bounding boxes. Each object prediction provides a vector $$p_i = \{l_i, x_i, y_i, w_i, h_i\}^T$$ where $l_i$ is the predicted object label, $x_i$ and $y_i$ the position of the top left corner of the predicted bounding boxes, and $w_i$ and $h_i$ its width and height.
The spatial information contained within $B$ consists of the information on scene geometry and objects' size. 
First, we use these predicted features to compute the Euclidean distances $\Delta_{i,j}$ between bounding boxes.
This Euclidean distance provides information on the geometric and spatial relationships between objects in the scene. This extracted information enables us to characterize the spatial arrangement between objects in the scene through geometric reasoning. 
The object size, expressed as the bounding boxes diagonals $d_i$ given by  $d_i = \sqrt{w_i^2 +h_i^2}$, represents an additional geometric attribute for the graph construction process.
This information, combined with the semantic information, provides a priori knowledge about the nature of the scene through the relationship between size and object type.
Indeed, a significant fluctuation in the size of objects of the same label contained in a scene induces a significant scene depth, increasing the probability of corresponding to an outdoor scene conversely.  
Therefore, our approach attempts to combine semantic and spatial reasoning into a coherent graph neural network.

\subsection{Construction of the Space-Semantic Graph}

The spatial semantic consistency of the detected object is integrated in a graph $\mathbb{G}$ composed of edges $\mathbb{E}$ and nodes $\mathbb{N}$ that represent the main scene information. The strength of our approach is in attributing both spatial and semantic information to nodes and spatial information to edges, which better describes the scene contents.

From a semantic point of view, only predicted labels $\mathbb{L}$ are considered to describe the node's attributes. 
The graph can be defined as $\mathbb{G} = \{\mathbb{V}, \mathbb{E} \}$, where nodes are expressed as $\mathbb{V} = \{ V_i:\{l_i, d_i\} | i \in \{ 1, ..., n \} \} $ integrating information from labels and bounding boxes.
Edges $\mathbb{E}$ describe the spatial distribution of objects through their Euclidean distance. 
However, unlike most other approaches, not all nodes are connected by edges to maintain the coherence of the spatial distribution of objects. 

We determine the nearest neighbours of a node $i$ by comparing its distance $\Delta_{i,j}$  for each node $j\not= i$.
The distance with its nearest node $\Delta i_{min}$ is expressed as follows:
\begin{equation}
    \Delta i_{min} = \min_{j \leqslant  n} \Delta_{i,j}
\end{equation}
\newline
\vspace{-2em}
\begin{figure}[h]
    \centering
    \includegraphics[width=0.44\textwidth]{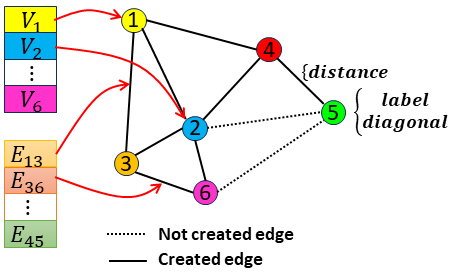}
    \vspace{-0.5em}
    \caption{Space-Semantic Graph construction.}
    \label{fig:graph}
    \vspace{-0.5em}
\end{figure}
\newline
This minimum distance is used to define the node connections of the graph. 
Only nodes at a distance close to $\Delta i_{min}$ up to a certain distance ratio $\beta$ are considered in the process of graph edges' construction. 
The construction of a space-semantic Graph corresponding to an observed scene is described in Algorithm \ref{algo1}.
\begin{algorithm}[!t]
\caption{Space-semantic Graph construction algorithm}\label{algo1}
\textbf{Input:} n predictions \{ Labels $L$,  Bounding boxes $B$\}, distance ratio: $\beta$\\
\textbf{Output:}  Graph $\mathbb{G}$
\begin{algorithmic}
\STATE Compute each object diagonal $d$
\STATE Compute each distance $\Delta_{i,j}$ between objects 
\STATE Determine distance $\Delta i_{min}$ of nearest object 
    \FOR{$i=1 $ to $n $}
            \STATE{Create Node $V_i$}
            \STATE{Assign to node $V_i \gets \{l_i, d_i\}$}
            \FOR{$j=1 \ to \ n $}
                \IF{$ j \neq i$}
                    \IF{$ \Delta_{i,j} \leqslant \Delta i_{min} + \beta \cdot \Delta i_{min}$}
                        \STATE {Create edge $E_{i,j}$}    
                        \STATE {Assign to edge $E_{i,j} \gets \Delta_{i,j}$}
                    \ENDIF
                \ENDIF
            \ENDFOR
    \ENDFOR
\STATE {$\mathbb{G} \gets \{ V , E \}$}
\end{algorithmic}
\end{algorithm}
As a result, nodes are only connected by edges to their nearest neighbours with $\mathbb{V} = \{ V_{i,j} : \{\Delta_{i,j}\} | i,j \in \{ 1, ..., n \} \} $. This nodes connectivity through specific edges is illustrated in Fig. \ref{fig:graph}.
The nodes connectivity is represented by an adjacency matrix $A$ representing existing edges. 
 $A$ is symmetric, but only some edges are created and have attributed values as shown in Tab.\ref{tab:matA}.

\subsection{Graph Convolutional Neural Network models}
Graph Convolutional Neural Networks (GCNN) models have been introduced to solve various problems in computer vision, especially for visual data classification, human action recognition, and face recognition, to name a few. In the following section, the most common  GCNN models are briefly described.
\begin{table}[!t]
\caption{Adjacency matrix $A$ of nodes.}
  \centering
  \begin{tabular}{c|c|c|c|c|c|c}
    \hline
    Edge & 1&2&3&4&5&6 \\
    \hline
    1& 0 & $\Delta_{1,2}$ & $\Delta_{1,3}$ & $\Delta_{1,4}$ & 0&0 \\
    2& $\Delta_{1,2}$ & 0 & $\Delta_{2,3}$ & $\Delta_{2,4}$ & 0&$\Delta_{2,6}$ \\
    3& $\Delta_{1,3}$ & $\Delta_{2,3}$ & 0 & 0 & 0&$\Delta_{3,6}$ \\
    4& $\Delta_{1,4}$ & $\Delta_{4,2}$ & 0 & 0 & $\Delta_{4,5}$&0 \\
    5& 0 & 0& 0 & $\Delta_{4,5}$ & 0&0 \\
    6& 0 & $\Delta_{6,2}$ & $\Delta_{6,3}$ & 0 & 0&0 \\
    \hline
  \end{tabular}
  \label{tab:matA}
  \vspace{-1.0em}
\end{table}

\subsubsection{Graph Convolutional Network model}

First, we propose the Graph Convolutional Network (GCN) model \cite{kipf2016semi} as a graph-based neural network model for our scene classification task. GCN is a multi-layer network that takes graphs $G$ composed of nodes $V$ and edges $E$.
A diagonal degree matrix $\Tilde{D}$ is expressed from the adjacency matrix $\Tilde{A}$ as follows:
\begin{equation}
    \Tilde{D}_{ii} = \sum_{j} \Tilde{A}_{ij}
\end{equation}
Where $\Tilde{A}$ is the adjacency matrix of the graph $G$.
The update state in the GCN layer is expressed as follows:
\begin{equation}
    h^{(k+1)} = \sigma \Bigl( \Tilde{D}^{-\frac{1}{2}} \Tilde{A} \Tilde{D}^{-\frac{1}{2}} h^{(k)} W^{(k)} \Bigr)  
\end{equation}
Where $\sigma$ is an activation layer, $h^{(k)}$ denotes activation of the $k^{th}$ layer, $W^{(k)}$ denotes the learnable weights in the layer $k$, and the first layer $h^{(0)}= V$.


\subsubsection{Graph Isomorphism Network  model}

Given that CNN architectures do not allow contextual relations to be extracted from a scene, we have used the Graph Isomorphism Network (GIN) model \cite{xu2018powerful}.
The GIN model is a variant of the GCN model \cite{kipf2016semi}, which enables the differentiation of graphs that are not isomorphic to each other. 
The computational model of the GIN update nodes in the graph convolutional layer is expressed as follows:
\begin{equation}
    h_v^{(k)}= MLP^{(k)} \Bigl((1+ \epsilon^{(k)}) \cdot  h_v^{(k-1)} + \sum_{u \in  \mathbb{N}(v) X)} h_u^{(k-1)}\Bigr) 
\end{equation}
Where $h_v^{(k)}$ denotes the feature representation of node $v$ of the $k^{th}$ hidden layer, $h_v^{(0)} = A_v$, MLP denotes Multilayer Perceptron, and $\epsilon^{(k)}$ is a learnable parameter. Here, $A_v$ is the attribute matrix of the $v^{th}$ node.

In addition, GIN has the advantage of concatenating information nodes across all layers of the model.
This readout function enables the graph classification by using individual node information.
The extracted intrinsic attributes of the scene are fed into the LogSoftmax layer to obtain the probability of the scene type for each class (indoor/outdoor), with the LogSoftmax function expressed as follows:
\begin{equation}
    LogSoftmax(x_i) = log \Bigl( \frac{exp(x_i)}{\sum_{j}exp(x_j)} \Bigr)
\end{equation}

\subsubsection{LAF Aggregation Module}
Pellegrini et al. \cite{pellegrini2020learning} propose a learnable aggregation function (LAF) corresponding to a generalized form of the $L_p$-norms, expressed as follows:
\begin{equation}
    L_{a,b}(x) := \Bigl( \sum_{i} x_{i}^{b} \Bigr)^{a} \ \ \ (a,b \geq 0)
\end{equation}
Where $x = \{ x_1, ..., x_N \}$ denotes a finite multi-set of real numbers, $L_{a,b}$ is invariant under the addition of zeros such as $L_{a,b}(x) = L_{a,b}(x \cup 0)$, with $0$ being a multi-set of zero cardinality.
The LAF layer can be defined as a restricted LAF function for sets $x \in [0,1]$:
\begin{equation}
    LAF(x) := \frac{\alpha L_{a,b}(x) + \beta L_{c,d}(1-x)}{\gamma L_{e,f}(x) + \delta L_{g,h}(1-x)}
\end{equation}
Where $a,b,...,h \geq 0$ are tunable parameters, and $\alpha, \beta, \gamma, \delta \in \mathbb{R}$.
Based on the Learnable Aggregation Function layer \cite{pellegrini2020learning}, we use the GINLAF model.

\section{Experiments}
\label{sec:expe}
In this section, all the experiments carried out to train, optimize, and evaluate the proposed method are provided and discussed.
The experiments were conducted on a computer with an Intel Xeon CPU and a NVIDIA Tesla T4 GPU.

\subsection{Dataset}
Training and evaluation were done on an MS-COCO \cite{lin2014microsoft} derived dataset providing scene type annotations, namely CD-COCO \cite{beghdadi2023cd}.
This dataset offers a major contribution by exploiting the advantages of the famous MS-COCO dataset, namely the very large number of annotated images containing object labels, bounding boxes and masks. 
In addition, the CD-COCO dataset provides the scene type, indoor/outdoor, of each images.
Consequently, it is not necessary to use an object detection model to provide inputs to train our GCNN scene classification model.

This database is divided into subsets of 72K, 9K, and 9K images for the training, validation, and test sets, respectively.
It is important to note that the images contained in the database are distorted, which reduces the effectiveness of the YOLACT model for object detection. 
\begin{figure}[!t]
\vspace{-0.5em}
    \centering
    \includegraphics[width=0.38\textwidth]{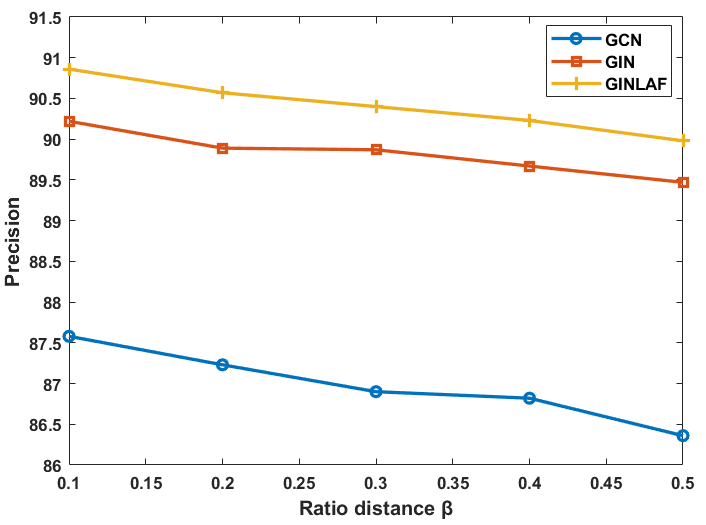}
    \caption{Model precision related to the beta distance ratio.}
    \label{fig:beta}
    \vspace{-1.3em}
\end{figure}

\subsection{Training}

The training was carried out on several GCN variant models to fully investigate the capabilities of our method. 
In our case study, we trained the GCN, GIN, and GINLAF models on multiple dimensions, which allowed us to conduct a comprehensive comparative study.
All models have been trained using the cross-entropy loss function to learn the parameter values.
The learning rate $Lr$, the weight decay $dd$ and the number of epochs are set to 0.001, $5e^{-4}$ and 10, respectively.
For the GCN and GIN models, the dimension of the hidden layers is set to 1024, while that of the GINLAF model is 32.
Finally, the batch size is set to 64 for models.

\subsection{Proposed GCNN-based model results}
First, tests were carried out using the database groundtruth (labels and bounding boxes) directly as input to our GCNN model. A detailed study was then carried out to assess the impact of the $\beta$ distance ratio on the performance of each model, as shown in Fig.\ref{fig:beta}.
The optimal $\beta$ value allows an increase of 1.22\%, 0.75\%, and 0.88\% for the GCN, GIN, and GINLAF models, respectively. 
This empirical study highlights the optimal performance for $\beta = 0.1$, which will be used for further experimentation.

Secondly, the effect of the number of object classes present in the observed scene on performance was evaluated. Indeed, the semantic content of a scene indirectly informs its intrinsic attributes. A scene composed of a large number of classes allows better extraction of the semantic relationships between objects, leading to a better classification, as shown in Tab.\ref{tab:training_result}. However, this ablation study does not consider the number of objects detected.
\begin{table}[!t]
\vspace{-1.5em}
\caption{Assessing method accuracy.}
  \centering
  \begin{tabular}{c|cccc}
    \hline
    & \multicolumn{4}{c}{Number of object classes}\\
    Method &3&4&5&6\\
    \hline
    GCN&87.5\% & 88.77\%&88.84\%&88.86\%\\
    GIN & 90.21\%&90.50\%&90.54\%&90.59\%\\
    GINLAF & \textbf{90.83\%}&\textbf{91.23\%}&\textbf{91.42\%}&\textbf{91.96\%}\\
  \end{tabular}
  \label{tab:training_result}
\end{table}
The GIN model seems to be the most robust to variations in the semantic content of the observed scene, while the GINLAF model offers the best performance.
A complete benchmark of various CNN and Vision Transformer (ViT) models has been performed to highlight the effectiveness and contribution of our method (see Tab.\ref{tab:benchmark}).
\newline
\begin{table}[!t]
\vspace{-0.5em}
\caption{Methods benchmarking.}
  \centering
  \begin{tabular}{p{1.8cm}|p{1.5cm}p{1.8cm}p{1.8cm}}
    \hline
    \multirow{2}*{Method} &\multirow{2}*{Accuracy $\uparrow$} &Parameter&Inference \\
    &&number $\downarrow$ & speed $\downarrow$ (ms)\\
    \hline
    GCN&88.9\% & 2,104,322&0.470\\
    GIN & 90.6\%&14,703,618&0.123\\
    GINLAF&92.0\% & \textbf{23,712}&\textbf{0.110}\\
    ResNet50 & 93.9\%&23,510,081&7.35\\
    DenseNet121&\textbf{94.6\% }& 7,978,856&22.9\\
    Basic ViT & 80.3\%&86,567,656&13.0\\
    \hline
  \end{tabular}
  \label{tab:benchmark}
  \vspace{-1.5em}
\end{table}

The accuracy, number of parameters and running time of each model have been determined to provide a complete and consistent comparison between the methods. 
This evaluation highlights the low computational cost of generating predictions using our method compared to CNN and ViT approaches for similar results (see Table.\ref{tab:benchmark}).
Indeed, our approach requires 100 times fewer parameters and inference speed is 66 times faster.
These performances justify the contribution of the approach, which guarantees good accuracy for a computational time suitable for its integration in an object detection model.

\subsection{YOLACT-GCNN framework results}

Our YOLACT-GCNN framework has been implemented and tested for the optimal $\beta$ parameter. 
In contrast to the previous study, using the outputs of the object detection model as inputs to the GCNN model induces an effect of the YOLACT \cite{bolya2019yolact} model performance on the classification performance. 
It should be noted that the object detection accuracy of the YOLACT model is reduced due to disturbances in the images contained in the CD-COCO database.
In addition, increasing the number of object classes detected does not result in a proportional increase in the number of objects detected.
The GCN model gains accuracy as the semantic content increases in terms of the number of different classes present in the observed scene.
As shown in Table \ref{tab:framework}, the GIN model behaves in the opposite way, regressing as the semantic content increases.
\begin{table}[h]
\caption{Framework performance evaluation.}
\vspace{-0.7em}
  \centering
  \begin{tabular}{c|cccc}
    & \multicolumn{4}{c}{Number of object classes}\\
    Method &3&4&5&6\\
    \hline
    GCN&76.4\% & 79.5\%&81.2\%&82.5\%\\
    GIN & 73.6\%&69.1\%&66.4\%&62.3\%\\
  \end{tabular}
  \label{tab:framework}
  \vspace{-0.7em}
\end{table}
\newline
Consequently, we can deduce that the GIN model takes greater advantage of the spatial distribution of objects in the scene for scene classification.
This assumption makes sense because of the GIN model's ability to differentiate between graphs that are not isomorphic.
Conversely, as the number of classes increases, the performance of the GCN model becomes more accurate due to its sensitive to semantic information, i.e. semantic relationships between objects.

\section{Conclusion}
\label{sec:conc}
Through this study, we have proposed a new lightweight GCNN-CNN models that allows us to overcome the limitations of CNN-based methods by incorporating semantic information, object proximity, and spatial information in the graph design. In addition, the proposed method is the first GCN model that can be used as an add-on to any object detection model. The experiment carried out on a dedicated dataset clearly demonstrates the efficacy of the proposed approach in indoor/outdoor scene classification. 

Our method achieves results that are very slightly lower than traditional CNN methods with a much lighter and easier-to-deploy network.
One limitation of the proposed solution is that it is sensitive to the reliability of the object detection module and the semantic nature of the scene complexity. Meaning that the more objects the scene contains the higher the accuracy achieved. It is worth noticing that the study could help design GCNN-based models for solving various computer vision problems. Finally, this new proposed solution could be considered as a starting point for GCNN-based non-satellite scene classification. 
\vspace{-0.5em}

{\small
\bibliographystyle{IEEEbib}
\bibliography{Euvip2024}
}

\end{document}